\documentclass[letterpaper, 10 pt, conference]{ieeeconf}  

\IEEEoverridecommandlockouts                              

\usepackage{graphicx}
\usepackage[table]{xcolor}  
\usepackage{booktabs}       
\usepackage{multirow}
\usepackage{threeparttable}
\newcommand{\up}{\ensuremath{{}\mkern2mu\uparrow}}
\newcommand{\down}{\ensuremath{{}\mkern2mu\downarrow}}
\usepackage{amsmath,amssymb}
\usepackage{cleveref}
\usepackage{csquotes}
\usepackage{float}
\usepackage{placeins}

\definecolor{mygreen}{RGB}{25, 100, 0}

\title{\LARGE \bf
MonoEM-GS: Monocular Expectation--Maximization Gaussian Splatting SLAM
}

\author{Evgenii Kruzhkov$^{1}$ and Sven Behnke$^{1}$
\thanks{$^{1}$All authors are with the Autonomous Intelligent Systems group, Computer
Science Institute VI – Intelligent Systems and Robotics – and the Center for
Robotics and the Lamarr Institute for Machine Learning and Artificial Intelligence, University of Bonn, Germany; 
        {\tt\small ekruzhkov@ais.uni-bonn.de}}%
\thanks{This work has been submitted to the IEEE for possible publication. Copyright may be transferred without notice, after which this version may no longer be accessible.}
}

\begin{document}

\maketitle
\thispagestyle{empty}
\pagestyle{empty}

\begin{abstract}
Feed-forward geometric foundation models can infer dense point clouds and camera motion directly from RGB streams, providing priors for monocular SLAM.
However, their predictions are often view-dependent and noisy: geometry can vary across viewpoints and under image transformations, and local metric properties may drift between frames.
We present MonoEM-GS, a monocular mapping pipeline that integrates such geometric predictions into a global \emph{Gaussian Splatting} representation while explicitly addressing these inconsistencies.
MonoEM-GS couples Gaussian Splatting with an \emph{Expectation--Maximization} formulation to stabilize geometry, and employs ICP-based alignment for monocular pose estimation. 
Beyond geometry, MonoEM-GS parameterizes Gaussians with multi-modal features, enabling in-place open-set segmentation and other downstream queries directly on the reconstructed map. 

We evaluate MonoEM-GS on 7-Scenes~\cite{glocker2013real}, TUM RGB-D~\cite{sturm12iros} and Replica~\cite{straub2019replica}, and compare against recent baselines.

\end{abstract}

\section{INTRODUCTION}
Simultaneous localization and mapping (SLAM) is a fundamental problem in robotics.
Monocular SLAM is particularly attractive because it requires only a single RGB camera, avoiding additional sensing hardware. 
Traditional monocular SLAM systems (e.g., ORB-SLAM2~\cite{murTRO2015}) rely primarily on visual features and geometric optimization, including feature matching, pose estimation, local bundle adjustment, and loop closure.
However, common limitations of traditional monocular SLAM are feature-poor regions and insufficient motion between frames.

Recently, feed-forward networks~\cite{wang2025vggt,wang2024dust3r,leroy2024mast3r,cabon2025must3r,wang2025pi3x,wang2025cut3r} that directly infer dense point clouds and camera poses from the frames have attracted significant attention for visual SLAM.
By extending the RGB input with additional geometric priors, new SLAM methods achieve dense monocular mapping in scenarios that are challenging for traditional pipelines.
Moreover, some models are trained to produce metric--scale priors.

However, a key limitation is that the predicted geometry can be noisy and inconsistent: the noise may vary across camera views and can even change for the same image under simple transformations such as vertical flipping. 
In addition, local metric properties (e.g., relative scale or segment lengths) can drift between consecutive frames. 
These inconsistencies pose a challenge for current approaches~\cite{maggio2025vggt,maggio2026vggt2,murai2025mast3r}, as they accumulate and degrade global consistency and smoothness of the reconstructed map.

In this work, we propose MonoEM-GS, a monocular approach that constructs the map by maximizing the predicted measurements' likelihood.
MonoEM-GS addresses the inconsistent predictions by merging them into the map in accordance with an incremental expectation--maximization (EM) pipeline and estimates the camera pose with respect to the current map~\cite{icp-color,icp-plane1,icp-planve2}.
We represent the map as a Gaussian Mixture Model (GMM) and employ Gaussian Splatting (GS)~\cite{kerbl20233dgs} to find challenging misalignments.
To the best of our knowledge, MonoEM-GS is the first approach that studies the combination of EM and GS within a single monocular SLAM system.

\begin{figure}[t!]
  \centering
  \includegraphics[
  width=\linewidth,
  height=0.8\textheight,
  keepaspectratio
]{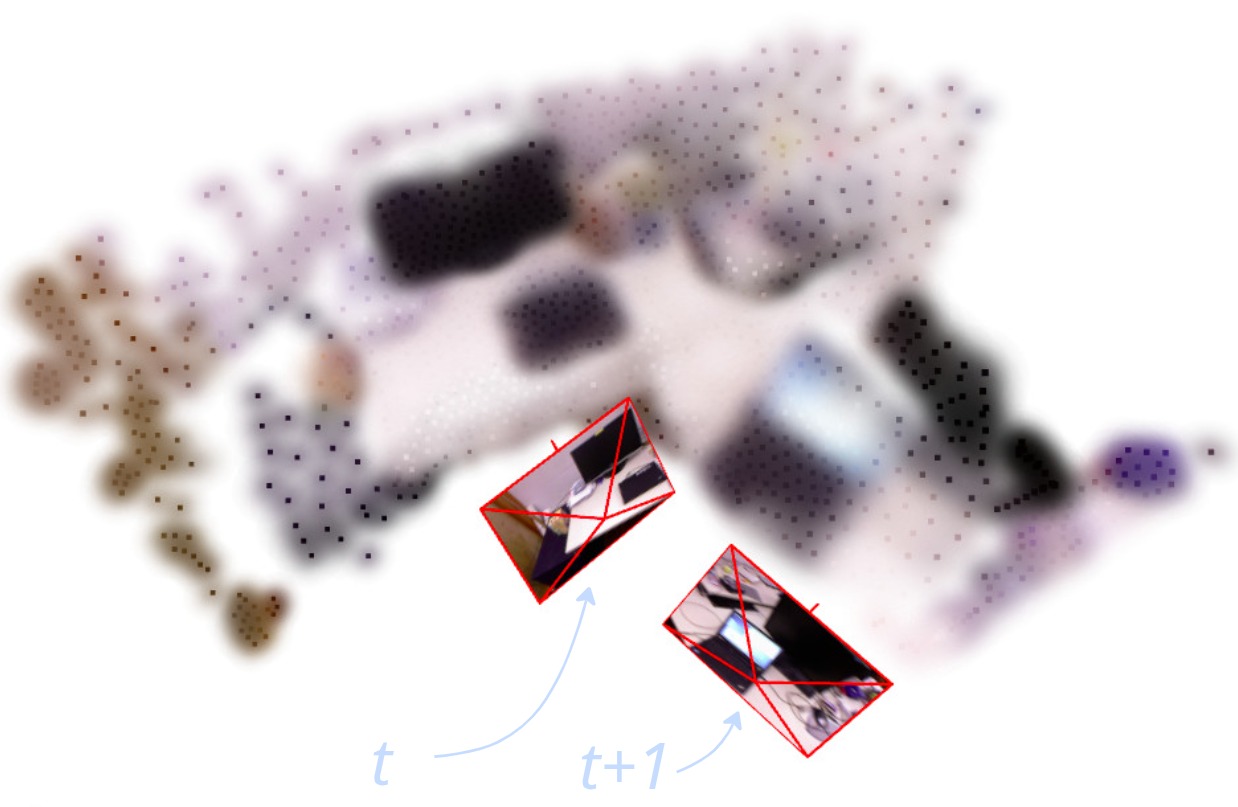}
  \caption{
  MonoEM-GS constructs a map in which each Gaussian aggregates measurements over time.
Despite contradictory point-cloud predictions at timestamps $t$ and $t{+}1$, the Gaussians remain consistent, with only a single update step needed to incorporate new observations.
  }
  \label{logo}
\end{figure}

Since geometric foundation models are integrated into the pipeline, they implicitly extract rich scene information. 
However, most SLAM systems do not explicitly attach this information to the resulting map, limiting what can be queried or computed directly from it.
MonoEM-GS goes beyond pure geometry by parameterizing Gaussians with multi-modal features, enabling a variety of downstream applications to operate directly on the map.
This is feasible because our Gaussian representation is sparser, allowing us to store high-dimensional features with no noticeable overhead.
\Cref{logo} demonstrates the Gaussians that were incrementally created by MonoEM-GS.

We summarize our contributions as follows:
\begin{itemize}
\item We present \textbf{MonoEM-GS}, a monocular mapping system that reconstructs point clouds and meshes with improved geometric consistency and supports in-place open-set segmentation directly on the map.
\item We show that \textbf{Gaussian Splatting} and \textbf{Expectation--Maximization} can be integrated within a single pipeline.
\item We demonstrate that \textbf{ICP}-based alignment can be used to obtain monocular pose estimates.
\end{itemize}

\section{Related Works}
The closest line of work to ours is learned-geometry monocular SLAM, where the measurements are predicted from monocular RGB by a feed-forward geometry model. 
MASt3R-SLAM~\cite{murai2025mast3r} builds monocular SLAM on two-view predicted reconstructions.
VGGT-SLAM~\cite{maggio2025vggt} extends this setting to an arbitrary number of frames and studies point-cloud alignment on the $\mathrm{SL}(4)$ manifold. 
VGGT2-SLAM~\cite{maggio2025vggt} further extends VGGT-SLAM~\cite{maggio2025vggt} and targets improved spatial alignment of internal submaps.
ViSTA-SLAM~\cite{zhang2025vista} also regresses geometry from two views using symmetric associations; unlike ~\cite{maggio2025vggt,maggio2026vggt2}, it assigns one pose-graph node per view. 
VGGT-Long~\cite{deng2025vggt-long} targets large-scale monocular SLAM while avoiding graph-based bundle adjustment.

As demonstrated by the aforementioned works, predicted geometry can serve as measurements for a SLAM system. 
However, such predictions are often noisy and view--dependent.
In contrast to prior works, we mitigate this by representing the map as a mixture model of parametrized Gaussians that are incrementally updated to increase the measurement likelihood. 
Overall, we target more geometrically consistent reconstructions by applying an EM-based~\cite{thrun2005bluebook} update to the predicted measurements and optimize the camera pose with respect to the map.



3D Gaussians in the form of Gaussian Splatting (GS)~\cite{kerbl20233dgs} have been successfully used for novel-view rendering within SLAM systems~\cite{monogs-dyson,sandstrom2025splat,keetha2024splatam}.
However, high reconstruction quality and low latency are typically achieved when additional geometric priors are available (e.g., depth sensors or LiDAR). 
HI-SLAM2~\cite{zhang2025hi} and SING3R-SLAM~\cite{li2025sing3r} follow this direction; in particular, SING3R-SLAM~\cite{li2025sing3r} uses predicted geometry as a prior.
In these methods, Gaussians are optimized primarily with novel-view synthesis objectives and updated via multiple per-frame optimization steps.
Although GS demonstrates strong performance in photorealistic rendering, it doesn't generally represent a probabilistic scene model and may overfit to the views.

In contrast, we combine EM with GS and treat the map as a Gaussian mixture. Our incremental EM-based map update aggregates incoming measurements in a single step per frame, while GS is used for improved spatial data association.
We don't optimize photorealistic rendering but employ rendering-based data association for the monocular predictions.  


A wide range of models (VGGT~\cite{wang2025vggt}, DUSt3R~\cite{wang2024dust3r}, MASt3R~\cite{leroy2024mast3r}, MUSt3R~\cite{cabon2025must3r}, Pi3-X~\cite{wang2025pi3x}, CUT3R~\cite{wang2025cut3r}, DA3~\cite{lin2025depthv3}, etc.) can generate geometric priors.
In this work, we use the MapAnything~\cite{keetha2025mapanything} framework to estimate the observable geometry from consecutive RGB frames.
We choose~\cite{keetha2025mapanything} because it is trained for metric-scale reconstruction and uses a DINO-based backbone, whose features are useful for downstream tasks.
Moreover, it provides a unified interface, enabling other 3D reconstruction models to be swapped in interchangeably.


Works \cite{maggio2025vggt,maggio2026vggt2,murai2025mast3r} produce dense point clouds, whereas our representation is sparse. 
As a result, we can additionally attach DINO~\cite{oquab2023dinov2,simeoni2025dinov3} features to each Gaussian with minimal memory overhead, enabling downstream tasks to operate directly on our reconstructions. 
Related multi-domain mapping has been demonstrated in OMCL~\cite{kruzhkov2026omcl}, OVO~\cite{martins2025open}, and RayFronts~\cite{alama2025rayfronts}.
Unlike our approach, OMCL~\cite{kruzhkov2026omcl} performs open-vocabulary Monte Carlo localization using map features, while OVO~\cite{martins2025open} relies on poses estimated by an external localization method.

\section{Method}\label{METHOD}

\begin{figure*}[t!]
\vspace{6pt}
  \centering
  \includegraphics[
  width=\linewidth,
  height=0.8\textheight,
  keepaspectratio
]{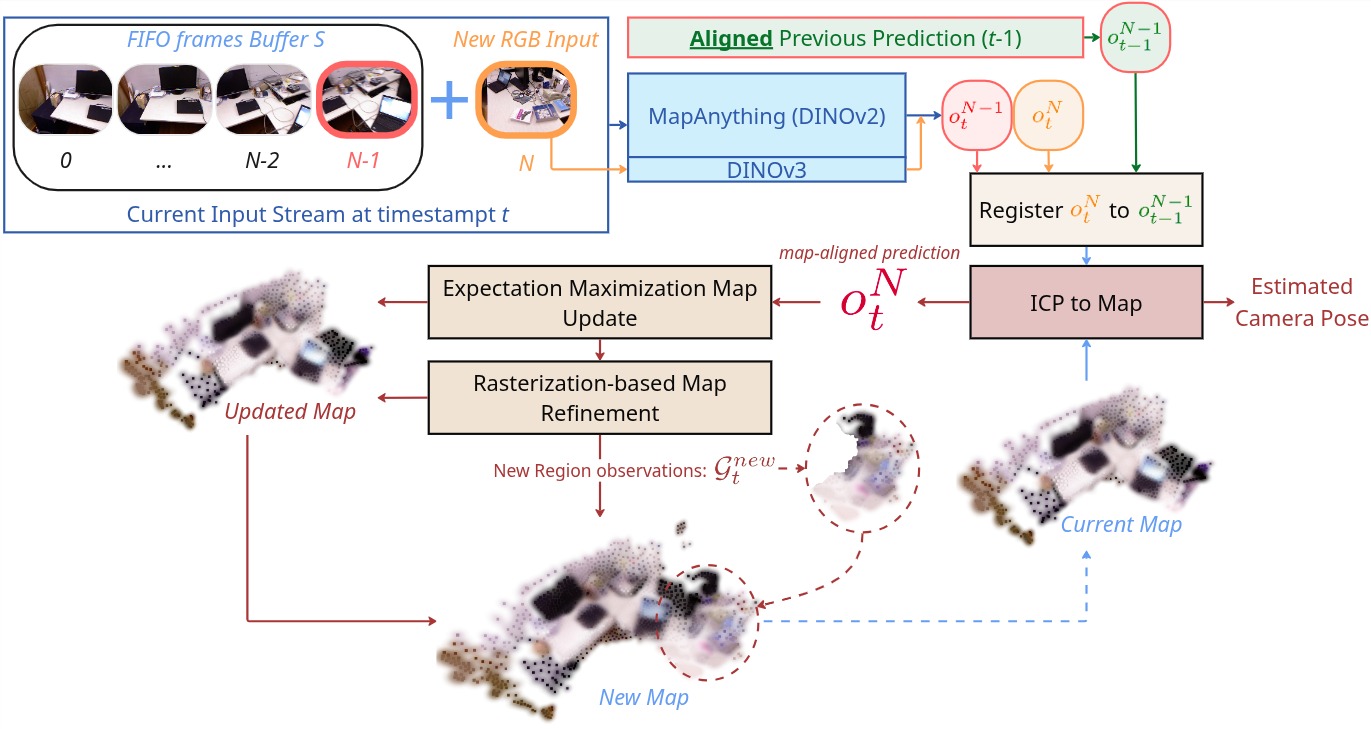}
  \caption{
  Single iteration of processing a new input image by MonoEM-GS.
  For each new RGB input, MapAnything~\cite{keetha2025mapanything} produces a dense point cloud prediction \textcolor{orange}{$o^{N}_{t}$}.
  We align it with the new prediction for frame N-1: \textcolor{red}{$o^{N-1}_{t}$} using ICP~\cite{icp-plane1,icp-planve2}.
  Since \textcolor{red}{$o^{N-1}_{t}$} and \textcolor{mygreen}{$o^{N-1}_{t-1}$} are created from the same frame at different timestamps, we employ the found correspondences between \textcolor{red}{$o^{N-1}_{t}$} and \textcolor{orange}{$o^{N}_{t}$} to register \textcolor{orange}{$o^{N}_{t}$} to \textcolor{mygreen}{$o^{N-1}_{t-1}$}.
  After the initial registration has been found, we perform colored ICP~\cite{icp-color}  alignment of \textcolor{orange}{$o^{N}_{t}$} to the map and estimate the camera pose with respect to the map.
  The aligned prediction \textcolor{purple}{$o^{N}_{t}$} is fused into the map and appended to the FIFO buffer.
  }\label{fig:pipeline}
\end{figure*}

We propose an approach for geometrically consistent monocular mapping, illustrated in~\Cref{fig:pipeline} for each new frame.
Our method represents the scene as a set of 3D Gaussians, which serve both as a mixture-model~\cite{thrun2005bluebook} map and as the primitives for Gaussian splatting rendering~\cite{kerbl20233dgs}.
The Gaussian parameters are accumulated across multiple consecutive observations predicted from the corresponding RGB views (\Cref{METHOD:ModelInference}).
We target monocular mapping with geometrically consistent reconstructions, reducing artifacts and shape ambiguities that can arise in SLAM pipelines that rely on feed-forward geometric predictions.
We assume calibrated inputs (known intrinsics), since calibration is inexpensive and improves the geometric predictions.

The proposed method starts with map initialization (\Cref{METHOD:Initialization}). 
For each new frame, it then localizes by aligning the new prediction to the current map (\Cref{METHOD:Localization}) and updates the map (\Cref{METHOD:Mapping}) by optimizing the Gaussian parameters to maximize the measurement likelihood.

\subsection{Model Inference}\label{METHOD:ModelInference}
We use the MapAnything~\cite{keetha2025mapanything} 3D reconstruction model to estimate the corresponding scene geometry from consecutive RGB frames, since it provides metric-scale reconstructions and a general-purpose feature backbone.
From each inference, we extract the camera pose, the predicted point cloud, the associated per-point colors and DINOv3~\cite{oquab2023dinov2,simeoni2025dinov3} features. We additionally compute per-frame surface normals for the point cloud.
Let $\mathcal{O}^N_t=\{o^n_t\}^N_{n=1}$ denote the set of inferred observations at timestamp $t$, where $N$ is the size of the set.

We found that the following heuristics improve the geometry predicted from RGB observations.

We run the model on a set of calibrated RGB images (known intrinsics) consisting of the latest image at time $t$; a fixed number of previous images stored in a FIFO buffer $\mathcal{S}$; and anchor image. The anchor is the first image of the dataset with known initial pose.

We provide the model with only the anchor pose. We do not provide poses for the other images in the input set, even if those poses were estimated earlier. This avoids over-constraining the model and improves its geometry predictions.


At each timestamp $t$, we integrate the previous frame, $t-1$, into the map, because its prediction improves when the input set includes the newer image at $t$. In other words, the newest image helps refine the geometry of the preceding one. Since the relative transformation between consecutive frames can be computed from the model prediction, we can temporarily estimate the pose of the frame $t$ relative to frame $t-1$ without introducing a one-frame pose output delay.

For simplicity, without loss of generality, we re-index time by assigning timestamp $t$ to the frame that is integrated into the map, and we refer to this frame as the \enquote{current frame}.

MapAnything~\cite{keetha2025mapanything} currently provides pre-trained weights for the DINOv2~\cite{oquab2023dinov2} model, while DINOv3~\cite{simeoni2025dinov3} offers more advanced capabilities for downstream applications.
Since a naive encoder swap is suboptimal, we currently extract DINOv3~\cite{oquab2023dinov2} features in parallel for the corresponding predictions.
However, future releases of~\cite{keetha2025mapanything} may be based on DINOv3~\cite{simeoni2025dinov3}.

\subsection{Initialization}\label{METHOD:Initialization}
To initialize the map, we fill an observation buffer $\mathcal{S}$ using a fixed input stride, and then construct the initial map from the merged predictions. 
For the construction, we cluster~\cite{douze2024faiss} the predicted 3D points into  $K$ clusters based on their coordinates and use the cluster centroids as Gaussian means.
The number of clusters $K$ is set automatically by dividing the total number of predicted points by a user-defined constant $\lambda$ and the number of inferred frames.

For each cluster $k$, we treat its centroid as the Gaussian mean $\boldsymbol{\mu}_k$ and estimate its covariance $\mathbf{\Sigma}_k$ from the coordinates of its $M$ nearest neighbors points $\mathrm{x}_m$.
Each Gaussian is additionally parameterized by a color $\mathbf{c_k}$, a unit surface normal $\mathbf{\bar{n}}_k$, and a DINOv3~\cite{simeoni2025dinov3} feature $\mathbf{f}_k$, computed from the same neighbors using Mahalanobis distance kernel weights $w_{km}=\exp(-\frac{1}{2}\mathrm{d}_{\Sigma_k}(\mu_k, \mathrm{x}_m))$:
\begin{equation}\label{eq:pts2gs}
\scalebox{1.}{$
\begin{aligned}
&\alpha_{km}=\frac{w_{km}}{\sum^{M}_{m=1}{w_{km}}},\;\;
\mathbf{\bar{n}}_k=\frac{\sum^{M}_{m=1}{\bar{n}_m\alpha_{km}}}{\|\sum^{M}_{m=1}{\bar{n}_m\alpha_{km}}\|_2},&\\
&\mathbf{f}_k=f_{k^*},\;\;\;\;\;\;\;
k^*=\underset{m}{\arg\max}\frac{f_m\sum^{M}_{m=1}{f_m\alpha_{km}}}{\|\sum^{M}_{m=1}{f_m\alpha_{km}}\|_2},&\\
&\mathbf{c}_k=\sum^{M}_{m=1}{\textrm{C}_m\alpha_{km}},
\end{aligned}
$}
\end{equation}

Our $\mathcal{M}$ consists of a set of parametrized Gaussians $\mathcal{M}=\{(\boldsymbol{\mu}_k, \mathbf{\Sigma}_k,\mathbf{c_k},\mathbf{\bar{n}}_k,\mathbf{f_k})\}^{K}_{k=1}$.

After initialization, the buffered frames are considered aligned with the map because it was directly constructed from their predictions.
For DINOv3~\cite{simeoni2025dinov3} features, we assign to each Gaussian the feature closest to the average feature of the neighboring points. 
The features are fixed after initialization and not updated together with the rest of the parameters.

\subsection{Localization}\label{METHOD:Localization}


We build the input stream from the buffered set $\mathcal{S}$ and the current image, predict the observations $\mathcal{O}_t$ of size $\lvert\mathcal{O}_t\rvert= \lvert\mathcal{S}\rvert+1$, and estimate the pose of the current frame with respect to the map $\mathcal{M}$ (following~\Cref{METHOD:ModelInference}, the captured frame is omitted).

\begin{figure}[t!]
  \vspace{6pt}
  \centering
  \includegraphics[
  width=\linewidth,
  height=0.8\textheight,
  keepaspectratio
]{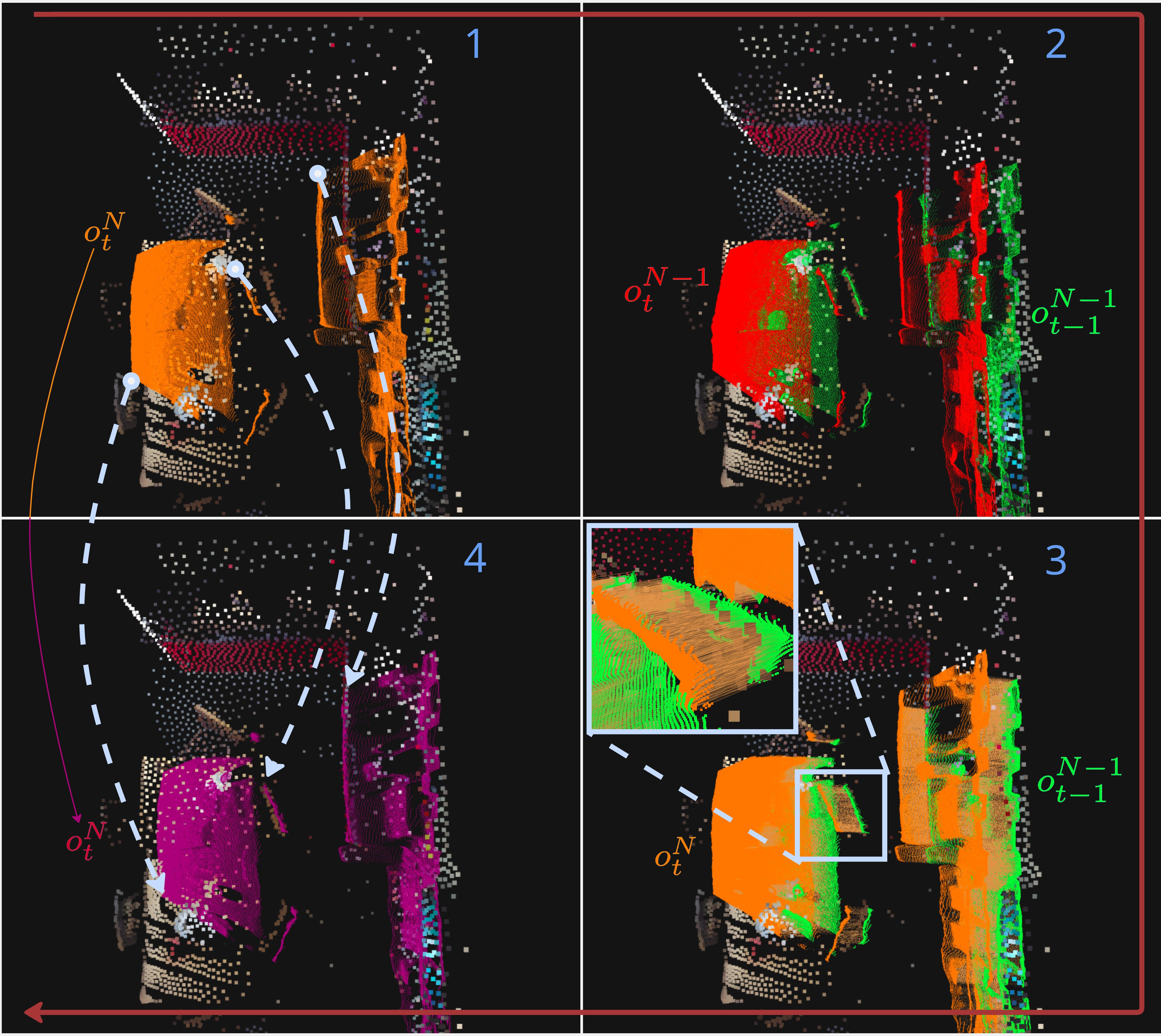}
  \caption{Views of the current prediction alignment with the map. 
  View 1: misaligned current prediction \textcolor{orange}{$o^{N}_{t}$};
  View 2: predictions \textcolor{red}{$o^{N-1}_{t}$} and \textcolor{mygreen}{$o^{N-1}_{t-1}$} for the same buffered image at different timestamps;
  View 3: the found associations between \textcolor{orange}{$o^{N}_{t}$} and \textcolor{mygreen}{$o^{N-1}_{t-1}$};
  View 4: current prediction \textcolor{purple}{$o^{N}_{t}$} aligned to the map.
  }
  \label{fig:alignment}
\end{figure}
The newly inferred prediction $o^{N}_t$ can be arbitrarily misaligned with $\mathcal{M}$.
For the buffered frames, however, we store their previous predictions, which are already aligned with the map (either because they were used during initialization or because they were aligned in earlier iterations).
Importantly, the last buffered image always appears in two consecutive inference steps.
Therefore, the predictions $o^{N-1}_{t-1}$ and $o^{N-1}_{t}$ are generated from the same input image and provide a one-to-one correspondence.


However, we require associations between predictions $o^{N}_{t}$ and $o^{N-1}_{t-1}$. 
To obtain them, we first compute closest-point correspondences between $o^{N}_{t}$ and $o^{N-1}_{t}$ using ICP~\cite{icp-plane1,icp-planve2} and thus derive the required point-wise associations between $o^{N}_{t}$ and $o^{N-1}_{t-1}$.
This yields a coarse registration of the current prediction $o^{N}_{t}$ to the map $\mathcal{M}$.

Next, we select a map region (submap) $\mathcal{M}^*_t$ for the precise alignment.
We employ the previously aligned point cloud of $o^{N}_{t-1}$ to compute an approximate bounding box of the currently observable region.
For efficiency on large maps, we query only the unique voxelized coordinates of Gaussians that fall inside this box, and then retrieve all Gaussians associated with the selected voxels. The retrieved Gaussians constitute the required submap.


The current frame pose is refined using colored ICP~\cite{icp-color} without scale estimation between $o^{N}_{t}$ and the submap.
As a result, the predicted point cloud for the current timestamp $t$ is tightly aligned with the map.
We depict all mentioned predictions in~\Cref{fig:alignment}


\subsection{Mapping}\label{METHOD:Mapping}

\textbf{Expectation--Maximization Map Update}. After aligning $o^{N}_{t}$ to the map, we integrate it into $\mathcal{M}^*_t$ using an incremental expectation--maximization (EM-based)~\cite{thrun2005bluebook} update of the Gaussian mixture model~\cite{thrun2005bluebook}, which increases the likelihood of the incoming measurements.
For every point $\mathbf{x}_{o^{N}_{t}}$ in $o^{N}_{t}$, we find its $K$ nearest Gaussians from $\mathcal{M}^*_t$.
Assuming the aligned measurements and the map have similar local geometry, we fuse into the map only those points whose average normal distances $d^{M\times K}_\mathrm{n}$ and Mahalanobis distance $d^{M\times K}_{\mathbf{\Sigma}_k}(\boldsymbol{\mu}_k, \mathbf{x}_{o^{N}_{t}})$ satisfy thresholds $\tau_\mathrm{n}$ and $\tau_\Sigma$: 
\begin{equation}
\begin{aligned}
    &d^{M\times K}_\mathrm{n}=\mathbf{\bar{n}}_k\bar{\mathrm{n}}_m, \quad\quad \frac{1}{K}\sum^K_k{[d^{M\times K}_\mathrm{n}]\geq\tau_\mathrm{n}} \\
    &\min_k(d^{M\times K}_{\mathbf{\Sigma}_k}(\boldsymbol{\mu}_k, \mathbf{x}_{o^{N}_{t}}))\leq\tau_\Sigma
\end{aligned}
\end{equation}

where $M$ is the total number of points in $o^{N}_{t}$.
The weakly explained predicted points that don't satisfy $\tau_n$ and $\tau_\Sigma$ are omitted at this step and will be fused during \enquote{Rasterization-based Refinement}.

For the rest of the points $\mathrm{M^*}$, we compute Gaussian--measurement responsibilities $r^{M^*\times K}$ (normalized measurement--likelihood weight) over the neighbors based on cosine similarity $d^{M^*\times K}_{\cos}$ of the DINOv3~\cite{simeoni2025dinov3} features, local normal consistency  $d^{M^*\times K}_\mathrm{n}$ and Mahalanobis distance $d^{M^*\times K}_{\mathbf{\Sigma}_k}$:

\begin{equation}\label{eq:resps}
\scalebox{.96}{$
\begin{aligned}
&p^{ik}_{det}=-\log\lvert\mathbf{\Sigma}_k\rvert \\
&p^{ik}=-\frac{1}{2}(d^{ik}_{\mathbf{\Sigma}_k})^2 + p^{ik}_{det} + \kappa_1 (d^{ik}_\mathrm{n}-1) + \kappa_2 (d^{ik}_{\cos}-1),\\
&r^{ik}=\frac{\exp{p^{ik}}}{\sum^{K}_{k=1}\exp{p^{ik}}},\;\; \pi_{i}=\max_k(p^{ik}) - \max_k(p^{ik}_{det})
\end{aligned}
$}
\end{equation}
for $i\in M^*, k \in K$, where $k$ is the k-th nearest Gaussian for point $i$; $\kappa_1$ and $\kappa_2$ are constants. 
We integrate into the map only the observations with $\pi_i\leq\tau_\pi$, i.e. only the observation for which the best estimated explanation is at least $\tau_\pi$ close to the best Gaussian.

\begin{figure*}[t!]
\vspace{6pt}
  \centering
  \includegraphics[
  width=\linewidth,
  height=0.8\textheight,
  keepaspectratio
]{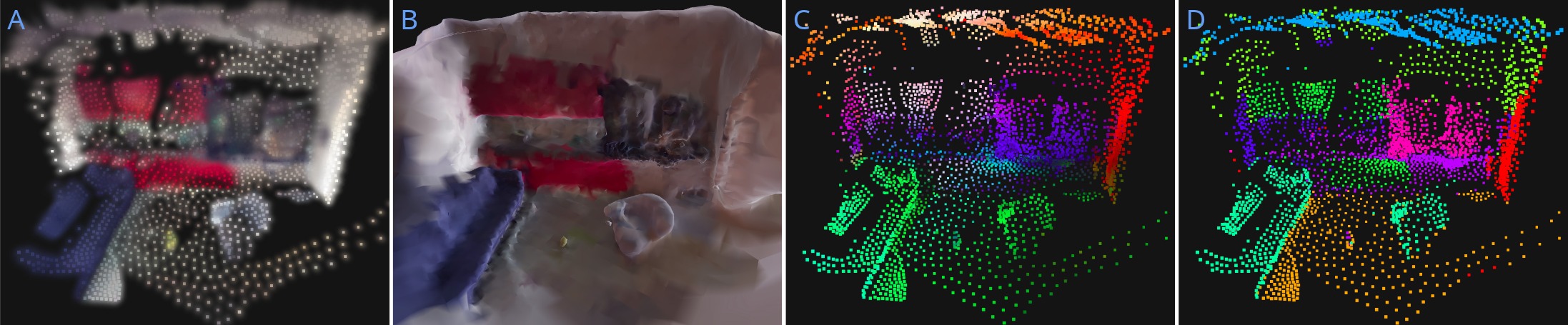}
  \caption{Multi-domain scene reconstructions produced by MonoEM-GS. \textcolor{cyan}{A}: Gaussian Splatting created by the proposed approach; \textcolor{cyan}{B}: Mesh reconstructed from Gaussian centers and normals; \textcolor{cyan}{C}: DINOv3~\cite{simeoni2025dinov3} feature map; \textcolor{cyan}{D}: Open-set 3D semantic map.}
  \label{fig:gs_mesh_features_sem}
\end{figure*}

Using the responsibilities as the weights, we estimate new parameters for the Gaussians  $\mathcal{M}^{*}_{t,new}$ from the current measurements $o^{N}_{t}$ (online M-step)~\cite{thrun2005bluebook}.
The final parameters are estimated~\cite{west1979updating} as a running average:
\begin{equation}\label{eq:M-step}
    \mathcal{M}^{*}_{t} \leftarrow (1-\alpha_t)\mathcal{M}^{*}_{t} + \alpha_t \mathcal{M}^{*}_{t,new},\;\alpha_t=\frac{\delta}{\alpha_{t-1}+\delta}
\end{equation}
where $\alpha_0=1$ and $\delta \in (0,1)$ is a per-Gaussian isotropy score that quantifies how symmetrically the assigned measurements support Gaussian $k$ around its center in the plane orthogonal to the normal $n_k$. 
In practice, for each tangential principal axis of Gaussian $k$, we split the assigned points into the positive and negative half-spaces according to their projection onto that axis, and compute $\delta_k=2\min(\sum_+ r^{ik}, \sum_-r^{ik})$.
Thus, $\delta_k$ is large when both sides of the principal axis receive similar total responsibility and small when the support is concentrated on only one side.
Note that the projection is used only to assign points to the positive and negative groups, and it uses the full responsibility values computed in~\Cref{eq:resps}.



\textbf{Rasterization-based Refinement}.
Many measurements could be omitted in the previous step either because they correspond to a new scene region or due to significant noise in the prediction.
To determine possible missed correspondences, we construct new parameterized Gaussians $\mathcal{G}_t=\{(\boldsymbol{\mu}_g, \mathbf{\Sigma}_g,\mathbf{c}_g,\mathbf{\bar{n}}_g,\mathbf{f}_g)\}$ from the predictions not fused by EM following~\Cref{eq:pts2gs}, and perform Gaussian splatting rendering~\cite{ye2025gsplat} for both $\mathcal{G}_t$ and $\mathcal{M}^*_t$, setting constant opacities and using the refined camera pose at timestamp $t$ and known intrinsics.

The rasterized $\mathcal{M}^{*R}_t$ produces expected depth, which we use to determine previously unmapped regions. 
A subset of $\mathcal{G}_t$ which was rasterized to pixels without valid depth is considered to be the observations of the new regions $\mathcal{G}^{new}_t$.
For the rest of the rendering $\mathcal{G}^R_t$, we estimate associations with $\mathcal{M}^{*R}_t$, finding the $K_{2\textrm{D}}$ closest candidates on the image plane from $\mathcal{M}^{*R}_t$ for each $\mathcal{G}^R_t$.
For each found correspondence, we estimate the 2D Mahalanobis distance on the image plane $d_{\Sigma_{\mathcal{M}}+\Sigma_{\mathcal{G}}}(\mathcal{M}^{*R}_t, \mathcal{G}^R_t)$ and cosine similarity $d_{cos}(\mathcal{M}^{*R}_t, \mathcal{G}^R_t)$.

We assume that correspondences with $d_{\Sigma_{\mathcal{M}}+\Sigma_{\mathcal{G}}}<\tau_{\Sigma 2\textrm{D}}$ and for which the rendered depth of the predictions is higher than that of the map belong to the same surface and can be merged. 
Those $\mathcal{G}^R_t$ which do not satisfy both $\tau_{\Sigma 2\textrm{D}}$ and the depth condition are considered new front observations and appended to $\mathcal{G}^{new}_t$.

The satisfied correspondences are merged with weights:
\begin{equation}    
\omega=\frac{\exp(-\frac{1}{2}d_{\Sigma_{\mathcal{M}}+\Sigma_{\mathcal{G}}})}{\sum_{K_{2\textrm{D}}}{\exp(-\frac{1}{2}d_{\Sigma_{\mathcal{M}}+\Sigma_{\mathcal{G}}})}}  d_{cos}(\mathcal{M}^{*R}_t, \mathcal{G}^R_t)
\end{equation}
where $\sum_{K_{2\textrm{D}}}$ is the sum over all $K_{2\textrm{D}}$ correspondences for each Gaussian in $\mathcal{G}^R_t$.

Next, we merge the components of $\mathcal{G}_t$ and $\mathcal{M}_t$. First, for each Gaussian in $\mathcal{M}_t$, we average the Gaussians of $\mathcal{G}_t$ associated with it with corresponding weights $\omega$.
The resulting averaged Gaussians $\bar{\mathcal{G}}_t$ have a one-to-one correspondence with $\mathcal{M}_t$ and, similarly to~\Cref{eq:M-step}, we update the map:
\begin{equation}
    \mathcal{M}^{*}_{t} \leftarrow (1-\gamma_t)\mathcal{M}^{*}_{t} + \gamma_t \bar{\mathcal{G}}_t,\;\gamma_t=\frac{\max_\mathcal{G}(\omega)}{\alpha_{t}+\max_\mathcal{G}(\omega)}
\end{equation}

Observations which were not fused with the map in both of the previous steps, $\mathcal{G}^{new}_t$, are appended to the map, extending it.
After the full localization and mapping step, the current frame with the aligned predictions is added to the FIFO buffer.

\begin{table}[t]
  \centering
    \caption{MonoEM-GS Parameters.}
  \label{tab:parameters}
   {
  \begin{tabular}{@{}lc|lc@{}}
    \toprule
    \textbf{Parameter} & \textbf{Value} & \textbf{Parameter} & \textbf{Value} \\
    \midrule
    Buffer size $\lvert\mathcal{S}\rvert$ & 10 & $\lambda$ & 128\\
    Normal distance $\tau_n$ & 0.5 & $\kappa_1$ & 5\\
    $\tau_\Sigma$ & 4.6 & $\kappa_2$ & 5\\
    Gaussians per point $K$ & 16 & $\tau_\pi$ (in $\log$-space)  & 8\\
    Gaussians per point $K_{2\textrm{D}}$ & 3 & $\tau_{\Sigma 2D}$ & 4.6\\
    \bottomrule
  \end{tabular}
  \normalsize }

\end{table}

\section{EXPERIMENTS}
Following MASt3R-SLAM~\cite{murai2025mast3r} and VGGT-SLAM~\cite{maggio2025vggt}, we evaluate on the TUM RGB-D~\cite{sturm12iros} and 7-Scenes~\cite{glocker2013real} datasets.
In addition, we evaluate the mapping on both datasets.
For pose evaluation, we report the RMSE of the absolute trajectory error (ATE) and the relative scale error of the estimated trajectory, computed using evo~\cite{grupp2017evo} with alignment.
For mapping quality, we report accuracy, completion, normal consistency, and F1-score at a 0.2 m threshold.
To obtain ground-truth scene geometry, we fuse the provided ground-truth depth images using the ground-truth poses for each dataset.

~\Cref{fig:gs_mesh_features_sem} illustrates the multiple out-of-the-box map representations supported by MonoEM-GS, highlighting its usability for a variety of robotics applications.
Following the evaluation protocol of OMCL~\cite{kruzhkov2026omcl} and OVO~\cite{martins2025open}, we report semantic segmentation performance on the Replica~\cite{straub2019replica} dataset using the dataset-provided ground-truth (GT) scene.

We specify parameters used by MonoEM-GS in~\Cref{tab:parameters}. All baseline methods are evaluated with their default parameters and, when supported, with calibrated intrinsics. All results are obtained on an NVIDIA RTX 3090 GPU with 24 GB of memory.
We process every every tenth image on all datasets.

\begin{table*}[t!]
\vspace{6pt}
\centering
\caption{Mapping and Localization Evaluation on 7-Scenes~\cite{glocker2013real} and TUM RGB-D~\cite{sturm12iros} Datasets.}\label{tab:slam_pose_rec}
\renewcommand{\arraystretch}{1.2}
\setlength{\tabcolsep}{5pt}
\begin{threeparttable}
\resizebox{\textwidth}{!}{
\begin{tabular}{lcccccccccccc}
\toprule
& \multicolumn{6}{c}{7-Scenes~\cite{glocker2013real}} & \multicolumn{6}{c}{TUM RGB-D~\cite{sturm12iros}} \\
\cmidrule(lr){2-7} \cmidrule(lr){8-13}
& \multicolumn{3}{c}{Meters [m]} & \multicolumn{3}{c}{Percentages [\%]}
& \multicolumn{3}{c}{Meters [m]} & \multicolumn{3}{c}{Percentages [\%]}\\
\cmidrule(lr){2-4} \cmidrule(lr){5-7} \cmidrule(lr){8-10} \cmidrule(lr){11-13}
\multirow{2}{*}{\centering \textbf{Methods}} &
\multirow{2}{*}{\centering \textbf{Acc.\down}}&
\multirow{2}{*}{\centering \textbf{Comp.\down}}&
\multirow{2}{7em}{\centering \textbf{Trajectory / ATE RMSE\down}} &
\multirow{2}{*}{\centering \textbf{F1\up}} &
\multirow{2}{*}{\centering \textbf{Normals\up}} &
\multirow{2}{*}{\centering \textbf{Scale\up}} 
&
\multirow{2}{*}{\centering \textbf{Acc.\down}}&
\multirow{2}{*}{\centering \textbf{Comp.\down}}&
\multirow{2}{7em}{\centering \textbf{Trajectory / ATE RMSE\down}} &
\multirow{2}{*}{\centering \textbf{F1\up}} &
\multirow{2}{*}{\centering \textbf{Normals\up}} &
\multirow{2}{*}{\centering \textbf{Scale\up}}
\\

\\
\cmidrule(lr){1-1} \cmidrule(lr){2-2} \cmidrule(lr){3-3} \cmidrule(lr){4-4} \cmidrule(lr){5-5} \cmidrule(lr){6-6} \cmidrule(lr){7-7} \cmidrule(lr){8-8} \cmidrule(lr){9-9} \cmidrule(lr){10-10} \cmidrule(lr){11-11} \cmidrule(lr){12-12} \cmidrule(lr){13-13}
VGGT-SLAM~\cite{maggio2025vggt}   & \cellcolor{green!35}\textbf{0.06} & 0.27 &  \cellcolor{green!15}0.07 & 70.7 & 34.0 & 47.2
& 0.15 & 0.27 & \cellcolor{yellow!35}0.05 & 63.2 & 46.2 & 74.8 \\
VGGT-SLAM2~\cite{maggio2026vggt2} & \cellcolor{yellow!35}0.09 & \cellcolor{green!35}\textbf{0.11} & \cellcolor{green!15}0.07 & \cellcolor{green!15}92.8 & \cellcolor{green!15}55.5 & 57.7 
& \cellcolor{yellow!35}0.11 & \cellcolor{yellow!35}0.21 & \cellcolor{green!15}0.04 & \cellcolor{yellow!35}79.6 & \cellcolor{green!15}62.2 & 65.8 \\
MASt3R-SLAM~\cite{murai2025mast3r}  & 0.11 & \cellcolor{yellow!35}0.16 & \cellcolor{green!35}\textbf{0.05} & \cellcolor{yellow!35}85.0 & \cellcolor{yellow!35}41.7 & \cellcolor{green!35}\textbf{88.4} 
& \cellcolor{green!15}0.08 & \cellcolor{green!15}0.19 & \cellcolor{green!35}\textbf{0.03} & \cellcolor{green!15}83.8 & 44.2 & \cellcolor{green!35}\textbf{86.1} \\

MapAnything~\cite{keetha2025mapanything} & 0.13 & \cellcolor{green!15}0.14 & 0.1 & 84.7 & 35.4 & \cellcolor{yellow!35}71.9 
& 0.15 & 0.24 & 0.13 & 65.9 & 38.9 & \cellcolor{yellow!35}75.4 \\
\textbf{MonoEM-GS (ours)} & \cellcolor{green!15}0.07 & \cellcolor{green!35}\textbf{0.10} & \cellcolor{yellow!35}0.08 & \cellcolor{green!35}\textbf{93.2} & \cellcolor{green!35}\textbf{59.8} & \cellcolor{green!15}77.3 
& \cellcolor{green!35}\textbf{0.06} & \cellcolor{green!35}\textbf{0.15} & 0.12 & \cellcolor{green!35}\textbf{86.8} & \cellcolor{green!35}\textbf{65.4} & \cellcolor{green!15}85.2\\ 
\bottomrule
\end{tabular}
}
\end{threeparttable}
\end{table*}
\subsection{Pose Estimation and Reconstruction}
For trajectory and reconstruction evaluation, we compare against similar RGB-based baselines and report the results in \Cref{tab:slam_pose_rec}. 
Since each Gaussian in our representation aggregates multiple observations, our map is naturally sparser.
To ensure a fair comparison of completeness, we densify the representation by rendering additional points (for each pixel using the dataset resolution) from the estimated poses.
For normal-consistency evaluation, we estimate normals for the baselines from their final point clouds, whereas our method estimates and stores normals during mapping, as described in~\Cref{METHOD:Initialization}.
Although MapAnything~\cite{keetha2025mapanything} is not a SLAM method, we include it in the evaluation because MonoEM-GS uses its predictions and MapAnything~\cite{keetha2025mapanything} can estimate camera poses from the input RGB sequence. 
For MapAnything~\cite{keetha2025mapanything}, we process 25 images sampled at equal temporal stride in a single batch to obtain its predictions, whereas the other methods operate sequentially on the stream.
We chose 25 images because this is the largest batch that fits in GPU memory on our hardware.


All evaluated approaches achieve similar performance, with only small differences between the top two methods. 
The proposed method is consistently ranked first or second across most metrics. 
It outperforms the MapAnything~\cite{keetha2025mapanything} results on all evaluated sequences and metrics.
Moreover, MonoEM-GS and MASt3R-SLAM~\cite{murai2025mast3r} maintain the most correct and consistent metric scale across all sequences.

7-Scenes~\cite{glocker2013real} contains small indoor scenes, whereas TUM RGB-D~\cite{sturm12iros} includes room-scale environments. 
Compared to optimization-based baselines, our pose estimates are less accurate in both environments; nevertheless, the final reconstruction-quality metrics remain comparable.

\begin{figure}[t]
  \centering
  \includegraphics[
  width=\linewidth,
  height=0.8\textheight,
  keepaspectratio
]{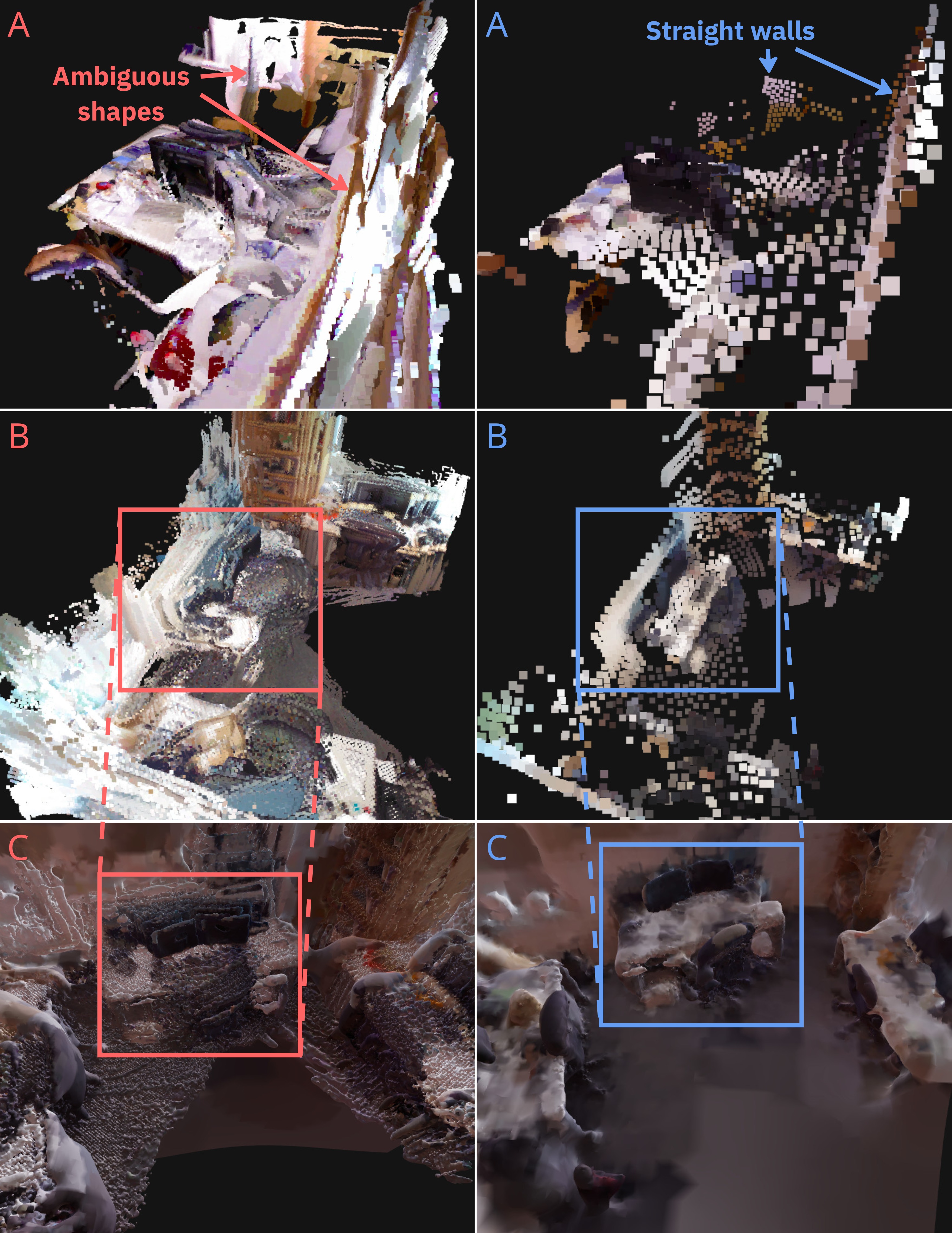}
  \caption{
  Qualitative comparison of VGGT-SLAM2~\cite{maggio2026vggt2} (\textcolor{red}{left}) and MonoEM-GS (\textcolor{cyan}{right}) maps. 
  \textcolor{red}{A}--\textcolor{cyan}{A}: VGGT-SLAM2~\cite{maggio2026vggt2} point cloud and MonoEM-GS Gaussian centers on the TUM RGB-D~\cite{sturm12iros} \enquote{desk2} sequence; 
  \textcolor{red}{B}--\textcolor{cyan}{B}: VGGT-SLAM2~\cite{maggio2026vggt2} point cloud and MonoEM-GS Gaussian centers on the 7-Scenes~\cite{glocker2013real} \enquote{office} sequence; 
  \textcolor{red}{C}--\textcolor{cyan}{C}: meshes reconstructed for both methods from the point clouds in \textcolor{red}{B} and \textcolor{cyan}{B}, respectively.
  }
  \label{walls_comparison}
\vspace{-0.1cm}
\end{figure}

However, the proposed approach yields more geometrically consistent reconstructions (\Cref{walls_comparison}). 
While the baselines recover many correct 3D points, they also accumulate erroneous predictions over time, which leads to ambiguous structures. 
In contrast, we achieve higher normal consistency, which is important for reliable mesh extraction.

\Cref{walls_comparison} compares point clouds and meshes produced by VGGT-SLAM2~\cite{maggio2026vggt2} and MonoEM-GS.
We chose VGGT-SLAM2~\cite{maggio2026vggt2} because it achieves the second-best normal-consistency score in \Cref{tab:slam_pose_rec} and is designed for improved measurement alignment. 
For our method, we use the normals estimated and stored during mapping. 
For VGGT-SLAM2~\cite{maggio2026vggt2}, we estimate normals from its final map, with manually tuned parameters to obtain the best result.

\begin{figure*}[t!]
\vspace{6pt}
  \centering
  \includegraphics[
  width=\linewidth,
  height=0.8\textheight,
  keepaspectratio
]{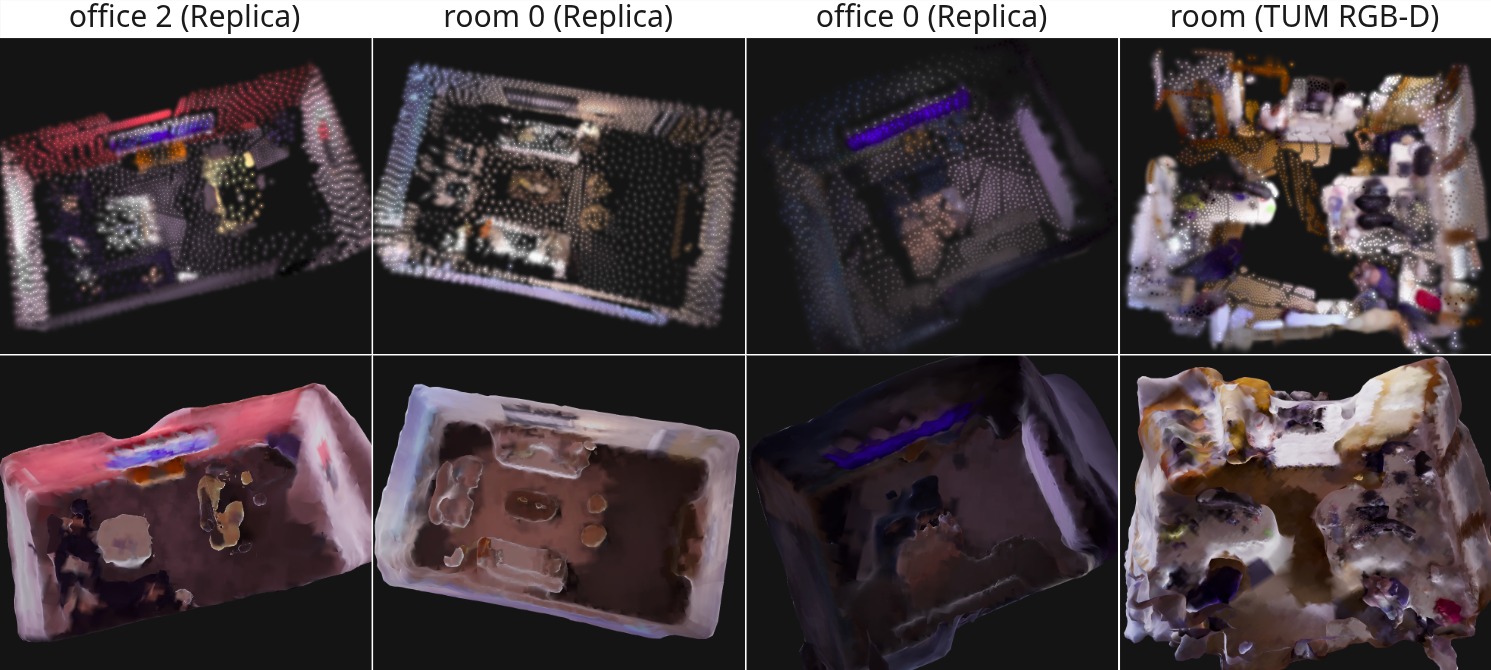}
  \caption{Reconstructed Gaussians (upper row) and corresponding meshes (bottom row) created from the Gaussian centers and normals.}
  \label{fig:demo_reconstruction}
\end{figure*}

As shown in \Cref{walls_comparison}, VGGT-SLAM2~\cite{maggio2026vggt2} produces a denser point cloud, but the geometry contains inconsistent and ambiguous shapes. 
In contrast, our map is sparser but exhibits clearer structure, with more unambiguous walls and surfaces. The mesh extracted from VGGT-SLAM2~\cite{maggio2026vggt2} contains substantial noise, whereas ours is cleaner.
Overall, our method achieves mapping-quality metrics comparable to the baselines while producing more consistent geometry.
An additional demonstration of our reconstruction results is provided in~\Cref{fig:demo_reconstruction}.

\begin{figure}[t!]
  \centering
  \includegraphics[
  width=\linewidth,
  height=0.8\textheight,
  keepaspectratio
]{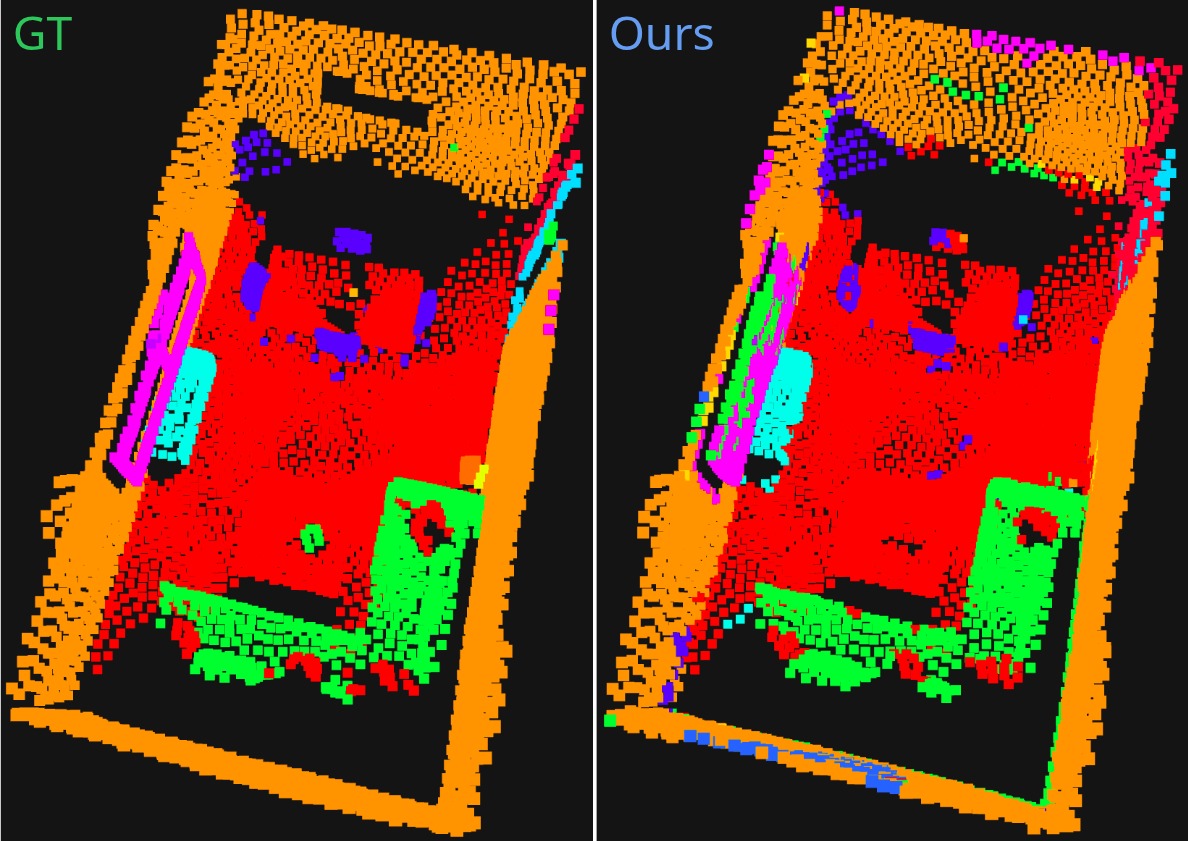}
  \caption{MonoEM-GS Gaussian centers colored by their predicted classes (\textcolor{cyan}{right}) and the nearest ground-truth points colored by their GT classes (\textcolor{mygreen}{left}).}
  \label{semantic_gt_comparison}
\end{figure}

\subsection{Semantic Segmentation}
We report average 3D semantic segmentation metrics on the photo-realistic Replica dataset~\cite{straub2019replica}, which contains room-scale environments. 
We directly use the DINOv3~\cite{simeoni2025dinov3} features stored in our map and classify them with the corresponding text encoder.
For each scene, we encode the dataset-provided class names with the text encoder and assign each Gaussian the class whose text embedding has the highest cosine similarity to its DINOv3~\cite{simeoni2025dinov3} feature.
Quantitative results are reported in \Cref{tab:semantic}.

\begin{table}[t!]
\caption{Replica 3D~\cite{straub2019replica} semantic segmentation evaluation.}\label{tab:semantic}
\resizebox{\columnwidth}{!}{%
\renewcommand{\arraystretch}{1.3}
\begin{tabular}{lcccc}
\toprule
\multicolumn{4}{c}{} & \multirow{2}{8em}{\centering \textbf{Trajectory /\\ ATE RMSE} {[}cm{]}}                                         \\
\textbf{Methods}         & \textbf{mIoU} {[}\%{]} & \textbf{f-mIoU} {[}\%{]} & \textbf{Acc} {[}\%{]} &  \\
\cmidrule(lr){1-1} \cmidrule(lr){2-2} \cmidrule(lr){3-3} \cmidrule(lr){4-4} \cmidrule(lr){5-5}
ConceptGraphs~\cite{gu2024conceptgraphs}   & 16.7          & 35.75           & 33.7         & N/A               \\ 
RayFronts~\cite{alama2025rayfronts}       & \cellcolor{yellow!35}27.7          & 43.47           & \cellcolor{green!15}54.5         & N/A               \\ 
OVO/ORB-SLAM2~\cite{martins2025open} & 27.1          & \cellcolor{green!15}60.2            & 39.1         & \cellcolor{green!35}\textbf{1.9}               \\ 
OML~\cite{kruzhkov2026omcl}             & \cellcolor{green!35}\textbf{32.1}          & \cellcolor{yellow!35}49.6            & \cellcolor{green!35}\textbf{56.2}         & \cellcolor{yellow!35}35                \\ 
\textbf{MonoEM-GS (ours)}            & \cellcolor{green!15}31.5          & \cellcolor{green!35}\textbf{63.2}            & \cellcolor{yellow!35}47.4         & \cellcolor{green!15}13.1              \\
\bottomrule
\end{tabular}%
}
\vspace{-0.3cm}
\end{table}

Note that our method estimates the trajectory as described in~\Cref{METHOD}, whereas other approaches either use ground-truth poses (N/A), rely on an independent localization system (OVO~\cite{martins2025open}/ORB-SLAM2~\cite{murTRO2015}), or decouple mapping from localization (OMCL~\cite{kruzhkov2026omcl}).

We achieve the best frequency-weighted mean IoU (f-mIoU) and the second-best mIoU. 
Although our mIoU is comparable to OMCL~\cite{kruzhkov2026omcl}, our f-mIoU is substantially higher because our predictions are more accurate on frequent classes but weaker on rare classes. 
In contrast, OMCL~\cite{kruzhkov2026omcl} attains a similar mIoU with a lower f-mIoU, suggesting more balanced performance across rare and frequent classes, but less consistent accuracy on the dominant classes.

Accuracy measures the fraction of points that are classified correctly and is influenced by rare classes. OMCL~\cite{kruzhkov2026omcl} achieves the highest accuracy, while our method and the second-best baseline correctly classify approximately half of the points.
~\Cref{semantic_gt_comparison} visualizes our predicted class for each Gaussian center and the ground-truth class of the nearest GT point to that center.

Our trajectory error on Replica~\cite{glocker2013real} (\Cref{tab:semantic}) is consistent with that on TUM RGB-D~\cite{sturm12iros} (\Cref{tab:slam_pose_rec}).
Although OVO~\cite{martins2025open} achieves better localization accuracy, it uses RGB-D input and relies on an external localization method. 
In contrast, we outperform OMCL~\cite{kruzhkov2026omcl}, which also performs localization using only monocular input.


\section{CONCLUSIONS}
In this paper, we introduced MonoEM-GS, a method for consistent monocular mapping. 
Overall, within a single monocular pipeline, MonoEM-GS reconstructs point clouds and meshes, stores multi-domain features, supports in-place open-set segmentation, and represents the scene as a set of 3D Gaussians whose centers and covariances define a probabilistic spatial density.
MonoEM-GS mitigates temporal measurement inconsistencies and increases the average measurement likelihood.
Compared to closely related methods, MonoEM-GS produces clearer geometry and smoother meshes.
However, MonoEM-GS is currently limited to small (room-scale) scenes due to the lack of loop closure. It may also integrate incorrect predicted measurements that lie significantly in front of the mapped surfaces.

\bibliographystyle{IEEEtran}
\bibliography{IEEEabrv,IEEEexample}

\addtolength{\textheight}{-12cm}   









\FloatBarrier
\end{document}